\documentclass[10pt,twocolumn,letterpaper]{article} 
\usepackage{avss}
\usepackage{times}
\usepackage{epsfig}
\usepackage{graphicx}
\usepackage{amsmath}
\usepackage{amssymb}
\usepackage[pagebackref=true,breaklinks=true,letterpaper=true,colorlinks,bookmarks=false]{hyperref}

\usepackage{lipsum}
\newcommand\blfootnote[1]{%
  \begingroup
  \renewcommand\thefootnote{}\footnote{#1}%
  \addtocounter{footnote}{-1}%
  \endgroup
}

\avssfinalcopy
\def\avssPaperID{59} 

\ifavssfinal\fi
\begin{document}
\renewcommand*\footnoterule{}

\title{Action Recognition with Domain Invariant Features of Skeleton Image}
\author{Han Chen \qquad Yifan Jiang \qquad Hanseok Ko\textsuperscript{*}\\
School of Electrical Engineering, Korea University\\
145 Anam-ro, Seongbuk-gu, Seoul, South Korea\\
{\tt\small\{jessicachan, yfjiang, hsko\}@korea.ac.kr}
}
\maketitle
\thispagestyle{empty}

\blfootnote{978-1-6654-3396-9/21/\$31.00 ©2021 IEEE}

\begin{abstract}
Due to the fast processing-speed and robustness it can achieve, skeleton-based action recognition has recently received the attention of the computer vision community. The recent Convolutional Neural Network (CNN)-based methods have shown commendable performance in learning spatio-temporal representations for skeleton sequence, which use skeleton image as input to a CNN. Since the CNN-based methods mainly encoding the temporal and skeleton joints simply as rows and columns, respectively, the latent correlation related to all joints may be lost caused by the 2D convolution. To solve this problem, we propose a novel CNN-based method with adversarial training for action recognition. We introduce a two-level domain adversarial learning to align the features of skeleton images from different view angles or subjects, respectively, thus further improve the generalization. We evaluated our proposed method on NTU RGB+D. It achieves competitive results compared with state-of-the-art methods and 2.4$\%$, 1.9$\%$ accuracy gain than the baseline for cross-subject and cross-view. 
\end{abstract}

\section{Introduction}
Human action recognition is one of the most popular research areas in computer vision, which has been widely used in surveillance systems, human-computer interaction, video understanding, etc. In the last few years, with the advance of 3D sensing techniques such as Microsoft Kinect, human action recognition utilizing 3D skeleton data arises a great deal of attention. Compared to RGB, skeleton data has the advantage of being computationally efficient due to smaller data sizes. Moreover, skeleton data are more robust to illumination, clustered background, and camera motion. 

With the increasing development and impressive performance of deep learning methods in most of the pattern recognition tasks, deep learning methods using Convolutional Neural Network (CNN), Recurrent Neural Network (RNN) and Graph Convolutional Network (GCN) with skeleton data also come into view. Some studies utilized the end-to-end learning based on RNN with Long-Short Term Memory (LSTM) to learn the temporal dynamics \cite{lee2017ensemble,si2019attention,han2020global,zhu2020exploring,ng2021multi,liu2017skeleton,song2017end}. Even if these methods present comparative results in skeleton-based action recognition due to their capability of modeling temporal sequence, they ignored the strong dependencies among the skeleton joints in the spatial domain. Recent studies have shown the superiority of CNN over RNN for this task \cite{ke2017new,li2017skeleton,le2018fine,caetano2019skelemotion,caetano2019skeleton,kim2017interpretable,liu2017enhanced}. Most of the CNN-based methods encoded the trajectories of skeleton joints into image space. They benefited from CNN's admirable ability to extract high-level spatial-temporal features of skeleton images. However, during convolutional operation, only the neighboring joints within the convolutional kernel will be considered. In other words, some latent correlation that is related to all joints may be neglected. Inspired by the fact that human 3D-skeleton is naturally a topological graph, the GCN has been adopted in the action recognition task recently for the reason that its effectiveness of representing the graph structure data \cite{li2018action,gao2019optimized,yoon2021predictively,li1802spatio,qin2020skeleton}. Most recent GCN-based methods assume that the thorough skeleton joints are available while not considering the incomplete case, which is common in real scenarios. 

\begin{figure*}
\centering
\includegraphics[width=16cm]{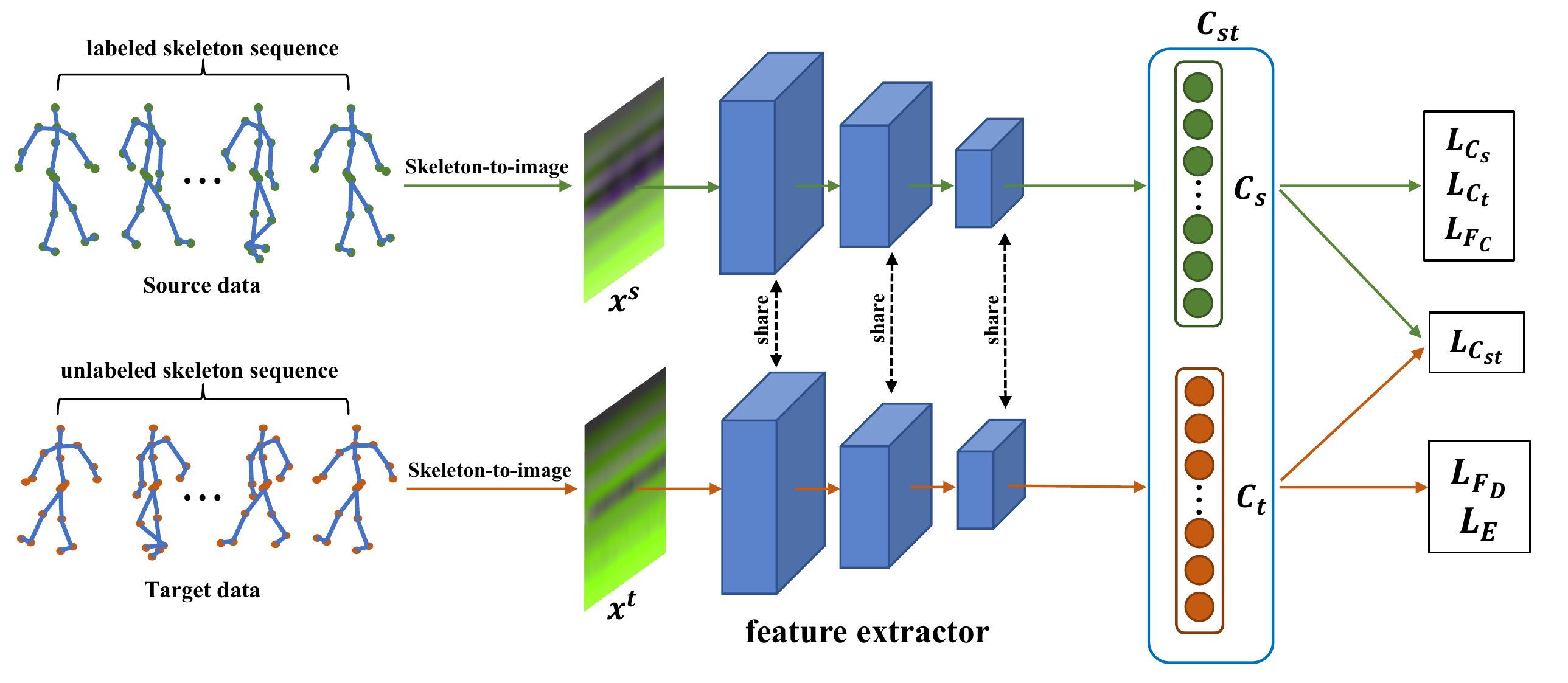}
\caption{An illustration of our proposed method architecture. We define the two domain inputs as source and target, where the source represents the labeled skeleton sequence of one view (or subject). The target represents the unlabeled skeleton sequence of another view (or subject). Our network includes a feature extractor, source task classifier ${C_s}$, target task classifier ${C_t}$ and a domain classifier ${C_{st}}$. Weights of all layers of feature extractor are shared for both source and target domain, and the ${C_{st}}$ shares neurons with ${C_s}$ and ${C_t}$. The green and orange colors illustrate the data flow of source and target, respectively. The right column shows a series of loss functions, please refer to the Section \ref{sec:2.2} and \ref{sec:2.3} for the definition details.}
\label{fig:overall}       
\end{figure*}


Unsupervised domain adaptation (UDA) aims to transfer a model learned on a labeled source domain to an unlabeled target domain. Recently, deep UDA has aroused great attention in image classification \cite{ganin2016domain,ghifary2014domain,long2015learning,long2017deep,sun2016deep,tzeng2017adversarial} and semantic segmentation \cite{huang2018domain,zhang2017curriculum,zou2018unsupervised,pan2020unsupervised}. There are several strategies proposed to gain better UDA performance. Some studies utilize maximum mean discrepancy (MMD) \cite{gretton2012kernel} to minimize differences between feature distributions \cite{ghifary2014domain,yan2017mind,haeusser2017associative}, its effect is limited by whether the distributions follow Gaussian distribution. Another strategy is self-training, which utilizes predictions from an ensemble model as pseudo-labels for unlabeled data to train the current model \cite{li2018disaster,zou2018unsupervised,spadotto2021unsupervised}. There is increasing interest in the use of adversarial training to achieve UDA \cite{vu2019advent,tzeng2017adversarial,pan2020unsupervised,jiang2021regressive,akkaya2021self}. This approach reduces the shift by forcing the features from different domains to fool the discriminator, thus leading to features from different domains exhibiting a similar distribution. For the task of action recognition, a universal and reliable system needs to be robust to different viewpoints while observing an action sequence. Therefore, several works have been proposed to utilized UDA to exploit disentangled universal representations from data of different views \cite{chen2019temporal,jamal2018deep,liu2017enhanced,li2018unsupervised,sigurdsson2018actor}. However, very few studies have attempted deep UDA for skeleton data, so this is still an open, challenging problem. 

Motivated by the success of the adversarial-training based UDA technique in image classification, we believe that CNN-based human action recognition from 3D skeleton data can benefit from this strategy. In this work, we propose a novel scheme to improve the CNN-based action recognition network, which utilizes adversarial training to drive the learned features of the intermediate network to be view-invariant or subject-invariant. We first encode the 3D skeleton sequence into a skeleton image, which served CNN for classification. For the labeled source skeleton images, the CNN network will take them as input and be trained in a supervised manner to have the classification ability. We assume that features of skeleton images from different views or subjects have a latent correlation. We propose utilizing the class-level and domain-level adversarial losses to enhance the invariant feature learning of skeleton images from different views or subjects.

The remainder of this paper is organized as follows. In Section \ref{Proposed Method}, we first give an overview of our proposed network and introduce three process phases. Experimental results and analysis are presented in Section \ref{sec:3}. Finally, we conclude this work in Section \ref{sec:4}.


\section{Proposed Method}
\label{Proposed Method}

This section presents our proposed method for action recognition with domain invariant features of skeleton images. In Figure \ref{fig:overall}, we present an overview of our proposed method. It consists of three phases: skeleton image representation, two domain task classifiers and high-level feature learning with a two-level domain adversarial strategy. Here, we define the two domains as source and target. The source means the labeled skeleton data of one view angle (or subject), the target represents the unlabeled skeleton data of another view angle (or subject). The aim is to utilize the unlabeled data to learn the domain invariant features of skeleton data, thus further improving the network's generalization ability to other views or subjects.
\subsection{Skeleton image representation}

As shown in Figure \ref{fig:overall}, given a skeleton sequence, we first map it to 2D image space. The X, Y, and Z coordinate information of the skeleton will be mapped to each channel of an RGB image, respectively. The rows represent different skeleton joints in the skeleton image, while the columns serve as the different frames. For the scenario where multiple subjects appear simultaneously, we concatenate the skeleton images of different subjects along the joint dimension. To suit the fixed input size of the 2D CNN, we further resize the skeleton image to a standard size. We denote the generated skeleton images from source and target as ${x^s}$ and ${x^t}$, respectively.

\subsection{Learning of source and target task classifiers}
\label{sec:2.2}
The proposed feature extractor follows the structure of ResNet-50 \cite{he2016deep} but excluding the last fully connected layer of it. There are two parallel classifiers ${C_s}$, ${C_t}$ for source skeleton image and target skeleton image respectively, each of them is based on a single fully connected layer followed by the softmax operation and contains ${K}$ neurons which equal to the class number of the actions. Besides, there is another domain classifier ${C_{st}}$, it shares neurons with ${C_s}$ and ${C_t}$.

The task classifier ${C_s}$ for the source domain is trained with the following cross-entropy loss over the labeled source skeleton images,

\begin{equation}
\label{eq1}
L_{C_s}=-\frac{1}{N_s}\sum\limits_{i=1}^{N_s}{y^s_i}\text{log}(p^s(x^s_i))
\end{equation}

\noindent where ${N_s}$ represents the number of source skeleton images, ${y^s_i}$ is the category label of the skeleton image, ${p^s(x^s_i)}$ is the output predicted probability of ${C_s}$.

To train the task classifier ${C_t}$ for the target domain, we utilize the labeled source skeleton image and update ${C_t}$ with the following loss function,

\begin{equation}
\label{eq2}
L_{C_t}=-\frac{1}{N_s}\sum\limits_{i=1}^{N_s}{y^s_i}\text{log}(p^t(x^s_i))
\end{equation}

\noindent where ${p^t(x^s_i)}$ is the output predicted probability of ${C_t}$.

To distinguish between the ${C_s}$ and ${C_t}$, we introduce the domain classifier ${C_{st}}$ in our network. It is achieved by applying the following loss,

\begin{equation}
\label{eq3}
L_{C_{st}}=-\frac{1}{N_s}\sum\limits_{i=1}^{N_s}\text{log}(\sum\limits_{k=1}^{K}p_k^{s}(x^s_i))
-\frac{1}{N_t}\sum\limits_{j=1}^{N_t}\text{log}(\sum\limits_{k=1}^{K}p_k^{t}(x^t_i))
\end{equation}

\noindent where ${N_t}$ represents the number of target skeleton images, ${\sum\limits_{k=1}^{K}p_k^{s}(x^s_i)}$ and ${\sum\limits_{k=1}^{K}p_k^{t}(x^t_i)}$ represent the probabilities of classifying the input skeleton image as the source and target respectively. By updating the ${C_{st}}$ with loss (\ref{eq3}), when we send the skeleton image of source domain to the network, the first term will be approximate to 1 and larger than the second term. Similarly, when we send the skeleton image of the target domain to the network, the second term will approximate to 1 and be larger than the first term. In this way, the ${C_{s}}$ and ${C_{t}}$ will be distinguishable.


\subsection{Learning of feature extractor through the two-level domain adversarial training}
\label{sec:2.3}

We utilize domain-level and class-level adversarial losses to align the feature distribution of source skeleton image and target skeleton image, respectively. These two losses are both achieved by adversarial learning, and through the two-level learning, the feature extractor tends to learn the domain invariant feature. 

We define the domain-level adversarial loss over the unlabeled target skeleton image as follows,

\begin{equation}
\label{eq4}
L_{F_D}=-\frac{1}{N_t}\sum\limits_{j=1}^{N_t}\text{log}(\sum\limits_{k=1}^{K}p_k^{s}(x^t_i))
-\frac{1}{N_t}\sum\limits_{j=1}^{N_t}\text{log}(\sum\limits_{k=1}^{K}p_k^{t}(x^t_i))
\end{equation}

\noindent where ${N_t}$ represents the number of source skeleton images, different from the loss (\ref{eq3}) used to update the classifier, the domain-level adversarial loss here is used to confuse the learned features of two domains to the maximum extent. That is, to make the feature extractor learn the common representation of source skeleton image and target skeleton image so that the classifier cannot distinguish which domain the feature map comes from.

The class-level adversarial loss is calculated over the labeled source skeleton image because we can get access to the class label of the source data. It is defined as the following function,

\begin{equation}
\label{eq5}
L_{F_C}=-\frac{1}{N_s}\sum\limits_{i=1}^{N_s}y_i^s\text{log}(p^{s}(x^s_i))
-\frac{1}{N_s}\sum\limits_{i=1}^{N_s}y_i^s\text{log}(p^{t}(x^s_i))
\end{equation}

To further improve the representation ability of the feature extractor, we also introduce the entropy minimization \cite{saito2019semi} to the updating stage. The principle is to encourage the low-density separation between classes by minimizing the entropy of the class-conditional distribution of target data \cite{grandvalet2005semi}. In our work, the loss is described in the follows,

\begin{equation}
\begin{split}
\label{eq6}
L_{E}=-\frac{1}{N_t}\sum\limits_{j=1}^{N_t}\sum\limits_{k=1}^{K}p_k^{s}(x^t_j)\text{log}(p_k^{s}(x^t_j)) \\
-\frac{1}{N_t}\sum\limits_{j=1}^{N_t}\sum\limits_{k=1}^{K}p_k^{t}(x^t_j)\text{log}(p_k^{t}(x^t_j))
\end{split}
\end{equation}


\subsection{Overall training objective function}

We define the overall loss function of our proposed approach as follows,

\begin{equation}
\begin{split}
\label{eq7}
\mathop{min}\limits_{C_s, C_t}(L_{C_s}+L_{C_t}+L_{C_{st}}) \\
\mathop{min}\limits_{F}(L_{F_C}+ \alpha(L_{F_D}+L_E))
\end{split}
\end{equation}

The first three terms are used to update the classifier layer, and the last three terms are for updating the feature extractor ${F}$. The ${\alpha}$ is the hyperparameter to balance different losses.

\section{Experiments}
\label{sec:3}
We evaluate the performance of our proposed method on the NTU RGB+D dataset \cite{shahroudy2016ntu}, we compare our proposed method with other state-of-the-art skeleton-based methods reported on the same dataset.

\subsection{Experimental settings}
\noindent \textbf{Dataset.} The NTU RGB+D dataset is the largest skeleton-based human action recognition dataset to date, captured with a Microsoft Kinect V2. It contains 60 action classes ranging from daily, health-related to two-person interactive actions with a total of 56,880 sequences. The skeleton data are composed of 25 joints. The actions are recorded at the same time by three cameras set in different view angles and performed by 40 different subjects. We follow the two standard evaluation setups for this dataset: Cross-Subject (CS) and Cross-View (CV), and take the training set and the test set as two different domain. Each skeleton image is resized to ${224\times224}$. We take the training set of the CS or CV setups as labeled source training data. As our method needs the unlabeled target data for training, so we randomly take 30\% of the original test set as the target training data and the left 70\% as our target test set.

\noindent \textbf{Implementation details.} We use PyTorch for implementation. All parameters are updated by the SGD optimizer. We set the batch size as 32 and the initial learning rate as 0.01. We follow the annealing strategy to update the learning rate. The hyperparameter ${\alpha}$ is updated by ${\alpha=\frac{2}{1+exp(-\gamma\times{p})}-1}$, where ${p}$ is the progress of the training epochs normalized to the range of 0 to 1, and ${\gamma}$ is a constant equals to 10. The structure of our network basically follows ResNet-50 \cite{he2016deep}. We change the last fully connected layer of the ResNet-50 to a total of ${2*}$class number neurons to construct the two task classifiers for the source domain and target domain. The other layers are treated as the feature extractor.

\subsection{Experimental results}
\noindent \textbf{Comparison with state-of-the-art.} In Table \ref{tab:1} we compare our proposed method with state-of-the-art skeleton-based methods on the NTU RGB+D dataset. The top-1 accuracy is chosen as the evaluation metric. The first group of methods are RNN-based methods, the second group of methods takes the CNN as the backbone, and the last group reports the results of GCN methods. The quantitative results show that our proposed method outperforms all previous RNN based and CNN-based methods in both cross-subject and cross-view scenarios, demonstrating that the learned feature representation is invariant to changes of view angle and subject. Our method also achieves comparable results with GCN-based methods, of which the computational complexity is typically over 15 GFLOPs for one skeleton sequence. Conversely, our method is composed of the lightweight CNN and appropriate losses. It just requires 3.8 GFLOPs for every skeleton sample. Please note that all state-of-the-art methods compared here are tested on the complete test set. However, our method achieves the domain invariant features learning by introducing data from the unlabeled target domain (30\% of the test set) for training. In order to achieve a fair comparison, we tested the effect of our baseline on 70\% test data and complete test data, respectively. We found that the difference between them is not significant, and the performance of the latter is even better than the former. Therefore, we can also prove that the results of the proposed method tested on 70\% test are comparable to the results of the 
SOTA methods above.

\begin{table}
\begin{center}
\begin{tabular}{l|c|c}
\hline
Method & CS & CV \\
\hline
GCA-LSTM \cite{liu2017skeleton} & 76.1 & 84.0 \\
STA-LSTM \cite{song2017end} & 73.4 & 81.2 \\
\hline
Clips + CNN + Concatenation \cite{ke2017new} & 77.1 & 81.1 \\
Clips + CNN + MTLN \cite{ke2017new} & 79.6 & 84.8 \\
Res-TCN \cite{kim2017interpretable} & 74.3 & 83.1 \\
Enhanced skeleton visualization \cite{liu2017enhanced} & 76.0 & 82.6 \\
F2CSkeleton \cite{le2018fine} & 79.6 & 84.6 \\
TSA \cite{caetano2019skelemotion} & 72.2 & 81.7 \\
TSRJI \cite{caetano2019skeleton} & 73.3 & 80.3 \\
\hline
A${^2}$GNN \cite{li2018action} & 72.7 & 82.8 \\
STGCK \cite{li1802spatio} & 74.9 & \textbf{86.3} \\
PA-GCN \cite{qin2020skeleton} & 80.4 & 82.7 \\
\hline
Our baseline (ResNet-50) on 70\% test & 78.6 & 83.9 \\
Our baseline (ResNet-50) on test & 78.8 & 84.1 \\
Ours & \textbf{81.0} & 85.8 \\
\hline
\end{tabular}
\end{center}
\caption{Comparison with state-of-the-art methods for action recognition on NTU RGB+D dataset in accuracy (\%).}
\label{tab:1}
\end{table}

\noindent \textbf{Effect of different loss terms in our method.} To demonstrate the effectiveness of our proposed method and different loss terms, we evaluate different variants of our method. All experiments are conducted on Cross-subject and Cross-view on NTU RGB+D dataset. First, we train ResNet-50 with the source skeleton image using the cross-entropy classification loss in a supervised manner and then test the trained network with the target test data (denoted as our baseline (ResNet-50)). Next, exclude the domain-level adversarial loss of our proposed method (denoted as Ours w/o ${L_{F_D}}$). Similarly, we exclude the entropy minimization loss (denoted as Ours w/o ${L_{E}}$) to examine its necessity. As shown in Table \ref{tab:2}, our proposed method improves the baseline accuracy by 2.4${\%}$ and 1.9${\%}$ for cross-subject and cross-view, respectively. This demonstrates the effectiveness of our proposed scheme for learning the domain invariant feature of the skeleton image. We also observe that in the cross-subject setup, there is a more significant accuracy drop when excluding ${L_{E}}$ than ${L_{F_D}}$, which represents that the entropy minimization contributes more than the domain-level adversarial loss in cross-subject. On the contrary, in the cross-view setup, the domain-level adversarial loss shows more importance than the other.

\begin{table}
\begin{center}
\begin{tabular}{l|c|c}
\hline
Method & CS (\%) & CV (\%) \\
\hline
Our baseline (ResNet-50) & 78.6 & 83.9 \\
Ours w/o ${L_{F_D}}$ & 80.6 & 84.9 \\
Ours w/o ${L_{E}}$ & 79.8 & 85.2 \\
\hline
Ours & 81.0 & 85.8 \\
\hline
\end{tabular}
\end{center}
\caption{Ablation study of different loss terms in our proposed method.}
\label{tab:2}
\end{table}

\begin{figure}[!t]
\centering
\includegraphics[width=8.5cm]{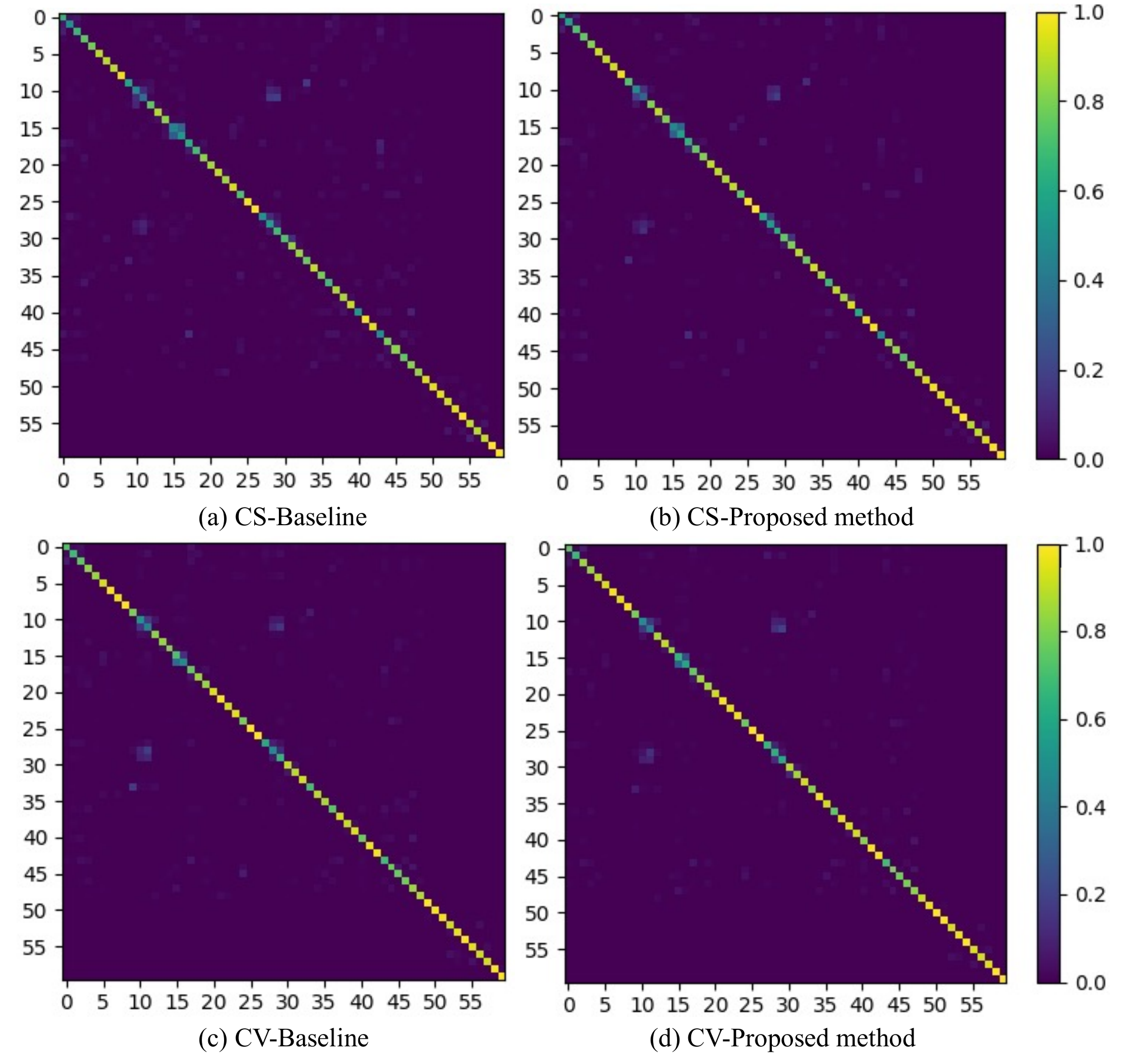}
\caption{Confusion matrix of the baseline (ResNet-50) and our proposed method on NTU RGB+D dataset across cross-subject (CS) and cross-view (CV) evaluation setups. X-axis (True class) and Y-axis (Predicted class) are associated through the indices of action classes.}
\label{confusion_matrix} 
\end{figure}

\noindent \textbf{Analysis of classification confusion matrix.} To further reveal the performance of our proposed method on each class, we show the confusion matrix on NTU RGB+D in Figure \ref{confusion_matrix}. The diagonal represents the correct classification for each action class. The non-diagonals show the misclassification results across different action classes. As shown in Figure \ref{confusion_matrix}, compared with the baseline, the confusion matrix of our proposed method is cleaner than baseline both in cross-subject and cross-view setups. In other words, our method can achieve more accurate prediction and fewer misclassification than baseline. The success of our method can be attributed to our two-level adversarial learning scheme, through which our network can learn the domain invariant features of skeleton images from different views or subjects. Besides, there is still some confusion in our results. For example, the action 11-reading tends to be classified as action 29-play with a phone. We attribute it to the fact that these two actions occur in similar motions and are prone to be confused when only using the skeleton images. In the future, our work will consider the RGB data as complementary information to the skeleton images so that to solve the above misclassification situation.

\section{Conclusions}
\label{sec:4}
In this paper, we proposed a novel action recognition network based on CNN and leverages unlabeled skeleton data from multiple views or subjects to learn view-invariant or subject-invariant feature representations of skeleton images. Our network learned the robust features for action recognition tasks by two-level domain adversarial learning strategy and entropy minimization. We trained our network on the NTU RGB+D dataset and demonstrated the effectiveness of our method on both cross-subject and cross-view setups. Experimental results showed that our proposed network outperforms baseline and state-of-the-art CNN-based methods. In the future, we will explore the fusion of the RGB modality into our network.

\section{Acknowledgement}
This work was supported by the Major Project of the Korea Institute of Civil Engineering and Building Technology (KICT) [grant number number 20210397-001].

{\small
\bibliographystyle{unsrt}
\bibliography{ref}
}
\end{document}